\pdfoutput=1

\documentclass[11pt]{article}

\usepackage{emnlp2021}

\usepackage{times}
\usepackage{latexsym}
\usepackage{color, graphicx}
\usepackage{eucal}
\usepackage[ruled,vlined]{algorithm2e}
\usepackage{enumitem}
\usepackage{multirow}
\setlist{noitemsep} 

\usepackage[T1]{fontenc}

\usepackage[utf8]{inputenc}

\usepackage{microtype}

\title{Effective Use of Graph Convolution Network and Contextual Sub-Tree for Commodity News Event Extraction}

\author{Meisin Lee, Lay-Ki Soon, Eu-Gene Siew \\
        Monash University \\
  \texttt{\{mei.lee, soon.layki, siew.eu-gene\}@monash.edu}}

\begin{document}
\maketitle
\begin{abstract}
Event extraction in commodity news is a less researched area as compared to generic event extraction. However, accurate event extraction from commodity news is useful in a broad range of applications such as understanding event chains and learning event-event relations, which can then be used for commodity price prediction. The events found in commodity news exhibit characteristics different from generic events, hence posing a unique challenge in event extraction using existing methods. 
This paper proposes an effective use of Graph Convolutional Networks (GCN) with a pruned dependency parse tree, termed contextual sub-tree, for better event extraction in commodity news. The event extraction model is trained using feature embeddings from ComBERT, a BERT-based masked language model that was produced through domain-adaptive pre-training on a commodity news corpus. Experimental results show the efficiency of the proposed solution, which outperforms existing methods with F1 scores as high as 0.90. Furthermore, our pre-trained language model outperforms GloVe by 23\%, and BERT and RoBERTa by 7\% in terms of argument roles classification. For the goal of reproducibility, the code and trained models are made publicly available~\footnote{\url{https://github.com/meisin/Commodity-News-Event-Extraction}}.
\end{abstract}

\section{Introduction} \label{sec:intro}
World events such as geo-political and macro-economic-related events have been shown to impact commodity prices in both short-term and long-term \cite{brandt2019macro}. Generally, events found in commodity news articles can be categorized into geo-political, macro-economic, supply-demand related, and commodity price movements. Commodity news is a valuable source of information to extract and mine for such events. Accurate event extraction is useful for many important downstream tasks, such as understanding event chains and learning event sequence, also known as scripts in \citep{schank2013scripts}, that can be used for accurate commodity price prediction ultimately. 

\begin{figure}[!t]
\centering
    \includegraphics[width=0.5\textwidth]{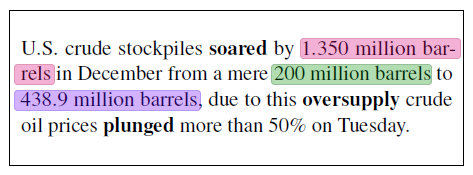}
    \caption{An example of a sentence from a piece of commodity news\protect\footnotemark, loaded with three events (trigger words are in bold): (1) Crude oil inventory increase, (2) Oversupply and (3) Crude Oil price decrease. Three event arguments that are of the same entity type are highlighted in color.}
    \label{fig:example1}
    \vspace{-0.5em}
\end{figure}

As defined in ACE (Automatic Content Extraction) Program\footnote{\href{https://www.ldc.upenn.edu/collaborations/past-projects/ace}{https://www.ldc.upenn.edu/collaborations/past-projects/ace}}, the event extraction task is made up of two subtasks: (1) event trigger extraction (identifying and classifying event triggers) and (2) event argument extraction (identifying arguments of event triggers and labeling their roles). In this work, we perform event extraction on the Commodity News dataset introduced in \cite{lee2021annotated}. Figure \ref{fig:example1} shows a sample sentence from this dataset. 

\footnotetext[3]{This example is used throughout this paper.}

To illustrate the task of event extraction, consider the example in Figure \ref{fig:example1} and the corresponding information tabulated in Table \ref{table:details_EE}. In the event trigger extraction sub-task, the model is trained to identify the trigger word: \textbf{soared} and classify the right event type: \textbf{movement\_up\_gain}, while in the event argument extraction sub-task, the model is trained to identify event arguments from a pool of entity mentions within the sentence and then label the argument roles each entity plays in relation to the identified event. 
\begin{table}[h!]   
    \centering \small
    \begin{tabular}{ |p{0.20\linewidth} | p{0.35\linewidth} | p{0.28\linewidth} |} \hline
    \textbf{Event} & \textbf{Entity Mention} & \textbf{Argument Role}\\ \hline
    Trigger: & U.S. & Supplier\\ \cline{2-3}
    \textbf{soared} & crude & Item\\ \cline{2-3}
    & stockpiles & Attribute\\ \cline{2-3}
     & 1.350 million barrels & Difference\\ \cline{2-3}
    Event type: & December & Reference point \\ \cline{2-3}
    \textbf{movement\_} & 200 million barrels & Initial Value\\ \cline{2-3}
    \textbf{up\_gain} & 438.9 million barrels & Final Value\\ \cline{2-3} 
     & more than 50\% & NONE \\ \hline
      \end{tabular}
      \caption{Event extraction of "crude oil inventory increase" event (first event in Figure \ref{fig:example1}).}
      \vspace{-1em}
    \label{table:details_EE}
\end{table}

\begin{figure*}[t]
\includegraphics[width=\textwidth]{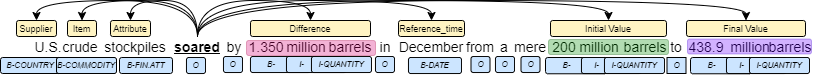}
\caption{The example in Figure \ref{fig:example1} is shown here with annotation provided in \cite{lee2021annotated}; trigger word is underlined and in bold, entity type for each entity mention is shown in blue below its respective word in BIO-tagging format while event argument role for each entity is shown in light yellow above the words; arches link argument to its trigger word.}
\label{fig:tagging}
\end{figure*}

The types of events found in commodity news are vastly different from generic events in ACE2005\footnote{ACE2005 is a multilingual Training Corpus developed under the ACE program}. Below is a list of unique characteristics of these events:
\begin{enumerate}[topsep=3pt]
    \item \textbf{Number intensity} - Numbers (e.g., price, difference, percentage of change) and dates (including date of the opening price, dates of closing price) are abundant in commodity news. These numerical data is critical in expressing financial information. Generic information extraction methods may not work well for numeric data, as it is clearly seen in \cite{saha2017bootstrapping} where the authors introduced the first Open \textit{numerical} relation extractor specifically to extract Open Information Extraction (IE) tuples that contain numbers or a quantity-unit phrase.
    \item \textbf{Arguments homogeneity} - Many arguments of the same entity type plays distinct roles in an event. Figure~\ref{fig:tagging} shows that \textit{1.350 million barrels}, \textit{200 million barrels}, \textit{438.0 million barrels} are tagged as \textit{QUANTITY} (see BIO-tagging at the bottom row).  However, all three arguments play a different role in relation to the event (see Argument roles at the top row). 
    \item \textbf{Undifferentiated event types without its arguments} - Consider the event triggered by \textbf{"soared"} in Figure \ref{fig:example1}, it is incomplete to identify the event (movement-up-gain) without  knowing  ``what” soared.  In commodity news,  possible events are `price soared’, `supply soared’, and `demand soared’.  To disambiguate events and represent them accurately, it is important to extract event arguments accurately as well.
\end{enumerate}


With these challenges in mind, this paper proposes a solution applying Graph Convolutional Networks (GCN) over contextual sub-tree for the task of event extraction. The contextual sub-tree is a dependency parse tree uniquely pruned that provides not just dependency path information but also off-path information. The off-path information adds more context to the existing dependency path between two nodes; hence it is termed contextual sub-tree. 

The contributions of this work are as follows: (1) We show that a domain-adaptive pre-trained Language Model, ComBERT, can yield promising performance over generic pre-trained language model when fine-tuned on event extraction tasks,
and (2) we propose an effective usage of a graph neural network, in the form of a GCN with contextual sub-tree that outperforms other existing approaches. The usefulness is particularly apparent in the subtask of event argument classification.

\section{Commodity News Dataset} \label{sec:dataset}
The dataset introduced in \cite{lee2021annotated} is a collection of annotated commodity news articles where its annotation is based on standards introduced by canonical programs such as ACE and TAC-KBP\footnote{\href{https://tac.nist.gov/2015/KBP}{https://tac.nist.gov/2015/KBP}}. Figure \ref{fig:tagging} captures in graphical form, the annotatations found in the dataset: (1) Entity Mentions - both named and nominal (entity type shown in blue under the sentence in BIO-tagging format); (2) Event Trigger Words (word underlined and in bold); and (3) Argument roles (labels are in light yellow above the sentence) with arches linking each argument to the event it belongs. This dataset consists of 21 entity types, 18 event types, and 19 argument role types (each event type has its own set of argument roles. Event types are defined based on Ravenpack's\footnote{RavenPack is an analytics provider for financial services. Among their products are finance and economic data. More information can be found on their page: https://www.ravenpack.com/} event taxonomy, out of which 18 event types are chosen. In terms of size, this dataset contains 8,850 entity mentions and 3,949 events. Table \ref{table:Distribution} shows the event type distribution. More information about the dataset, including argument roles, entity types, and example event types are found in Appendix \ref{app:Dataset}.

\begin{table}[h!]   
    \centering \small
    \begin{tabular}{ |p{0.50\linewidth} |r | c|}  \hline
    \textbf{Event type} & \textbf{Type ratio} & \textbf{\# sentence} \\ \hline
    1. Cause-movement-down-loss & 13.35\% & 524  \\ \hline
    2. Cause-movement-up-gain & 2.23\% & 88 \\ \hline
    3. Civil-unrest & 2.53\% & 100\\ \hline
    4. Crisis & 0.76\% & 30 \\ \hline
    5. Embargo & 3.75\% & 148 \\ \hline
    6. Geopolitical-tension & 1.70\% & 67 \\ \hline
    7. Grow-strong & 6.03\% & 238 \\ \hline
    8. Movement-down-loss & 22.69\% & 896 \\ \hline
    9. Movement-flat & 1.52\% & 60 \\ \hline
    10. Movement-up-gain & 22.13\% & 874 \\ \hline
    11. Negative-sentiment & 4.79\% & 189 \\ \hline
    12. Oversupply & 2.63\% & 104 \\ \hline
    13. Position-high & 3.82\% & 151 \\ \hline
    14. Position-low & 3.11\% & 123 \\ \hline
    15. Prohibiting & 1.06\% & 42 \\ \hline
    16. Shortage & 1.04\% & 41 \\ \hline
    17. Slow-weak & 5.47\% & 216 \\ \hline
    18. Trade-tensions & 1.39\% & 55 \\ \hline \hline
    Total & & 3949 \\ \hline
    \end{tabular}
    \caption{Event type distribution and sentence level counts}
    \label{table:Distribution}
\end{table}

\section{Related Work} \label{sec:relatedWork}
\paragraph{Event Extraction in Finance/Economics.} Generic event extraction-related work is covered in detail in the survey paper \cite{8918013} and in \cite{hogenboom2016survey}. Here we focus specifically on event extraction within the domain of finance and economics. Most of the event extraction tasks in this domain focuses on extracting company-related events. Here is a summary of related methods in recent literature: 
\begin{enumerate}[topsep=3pt]
    \item Rule-based approach: authors in \cite{malik2011accurate} introduced statistical classifiers aided by rules, while authors in \cite{hogenboom2013semantics} used rule-sets and domain ontology knowledge-bases together with other semantically-enabled components;
    \item Usage of external resources: Semantic Frame was in \cite{xie-etal-2013-semantic} and Wikipedia for weak supervision in \cite{ein-dor-etal-2019-financial};
    \item Open IE (Open Information Extraction): authors in \cite{ding2014using} extracted events from news headline via Open IE; while in \cite{saha2017bootstrapping}, authors introduced an extension of Open IE to extract numerical arguments in each Open IE tuple.
    \item Deep learning approach: \cite{yang-etal-2018-dcfee} proposed a deep learning approach to extract financial events from Chinese text.
\end{enumerate}

Although company financial events and commodity news fall under the same domain,  and both may involve numerical data as event arguments, existing methods for company financial event extractions are rather limited. For example, the solution in \cite{saha2017bootstrapping} caters for extracting only one numerical argument for each Open IE tuple. In comparison, the task of extracting company financial information in \cite{yang-etal-2018-dcfee} is the closest match to this work in terms of extracting numerical data as event arguments. However, it is a solution for Chinese text and focuses on document-level event extraction.

\paragraph{Graph Convolutional Networks.} The usage of Graph Convolutional Networks (GCN) coupled with syntactic information from dependency parse tree has been used for event extraction in \cite{nguyen2018graph} and in \cite{liu-etal-2018-jointly}. In \cite{nguyen2018graph}, the authors proposed using GCN over syntactic dependency graphs of sentences to produce non-consecutive \textit{k}-grams as an effective mechanism to link words to their informative content directly for event detection. Authors in \cite{liu-etal-2018-jointly} on the other hand, used attention-based GCN to model graph information to extract multiple event triggers and arguments jointly. Their proposed solution, Joint Multiple Events Extraction (JMEE) framework, focuses on modeling the association between events to enhance the accuracy of event extraction. Both these solutions use the shortest dependency path. 

Apart from event extraction, GCN has been used successfully for relation extraction in \cite{zhang-etal-2018-graph}. Instead of obtaining tokens strictly from the shortest dependency path, authors in \cite{zhang-etal-2018-graph} made modifications to produce pruned a sub-dependency tree to include off-path information as well such as negation cue words. 
Among the related work listed here, the one that is closest to our work in terms of the task (event extraction) and scope (sentence level) is JMEE by \cite{liu-etal-2018-jointly}.


\section{Proposed Solution} \label{sec:Proposed_solution}
In reference to the list of unique characteristics of events found in commodity news (listed in Section~\ref{sec:intro}), this paper proposes a solution using Graph Convolutional Networks (GCN) with contextual sub-tree for effective event extraction in commodity news. 
\begin{figure*}[t]
\includegraphics[width=\textwidth]{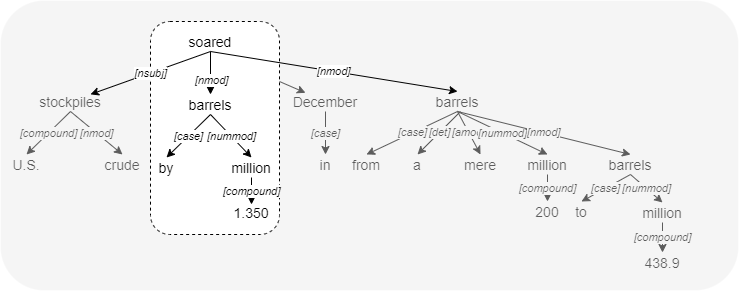}
\caption{The dependency parse tree of the example sentence.}
\label{fig:sub_parse_tree}
\end{figure*}

Syntactic dependency graphs represent sentences as directed trees with head-modifier dependency arcs between related words. Each word in such graphs is surrounded by its direct syntactic governor and dependent words (the neighbors), over which convolution operations can be performed on the most relevant words and avoid the modeling of unrelated words. Rather than using the full dependency parse tree, we propose to use a uniquely pruned dependency tree that is made up of the shortest path between two nodes (in our case - the \textbf{trigger candidate} and \textbf{entity mention}) and additional \textbf{off-path} nodes. The off-path nodes are included to provide additional contextual information. The resulting tree is termed \textbf{contextual sub-tree}. 

\subsection{Contextual Sub-tree}
The dependency tree for the example sentence is shown in Figure \ref{fig:sub_parse_tree} with one of the many candidate contextual sub-trees highlighted. Inspired by \cite{zhang-etal-2018-graph}, we prune the dependency tree to obtain the sub-tree rooted at the Least Common Ancestor (LCA) between the trigger candidate and the entity mention candidate while also contains off-path nodes. These off-path nodes provide additional and crucial contexts that enable better results in argument role classification. Off-path information is made up of tokens that are up to distance $DIST$ away from the dependency path. Algorithm~\ref{algo} shows the steps of how to build the contextual sub-tree. As shown in \cite{zhang-etal-2018-graph}, $DIST = 1$ achieves the best balance between including contextual information and keeping irrelevant ones out of the resulting sub-tree as much as possible.
    
\begin{algorithm}[ht!]
\SetAlgoLined
\KwResult{sub-tree structure}
 convert head indexes to tree object\;
  \eIf{DIST $<$ 0}{
   build the whole tree\;
   }
   {find all ancestor nodes of trigger\;
   find all ancestor nodes of entity\;
   find lowest common ancestor\;
   generate PathNodes (common nodes between trigger \& entity)\;
   insert more nodes based on DIST away from PathNodes\;
   }
 \caption{Build sub-parse tree from dependency head indexes}
 \label{algo}
\end{algorithm}

\begin{figure}[ht]
\includegraphics[width=0.4\textwidth]{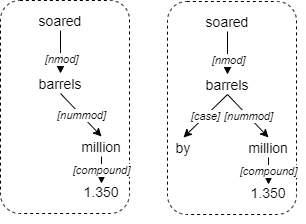}
\caption{Left: Sub-tree with shortest path, Right: Contextual sub-tree with off-path information}
\label{fig:parse_tree_combined}
\end{figure}

The usage of contextual sub-tree is targeted at \textbf{argument role classification}, the subtask within event extraction that classifies the argument role each entity plays in an event. Entity candidates are classified into one of the 19 argument roles. Figure~\ref{fig:parse_tree_combined} (Left) shows the sub-tree with the LCA path between event trigger and argument, while Figure~\ref{fig:parse_tree_combined} (Right) shows a slightly different sub-tree that not only contains the LCA path but also ``off-path'' information. 
    
Below are three examples of contextual sub-trees (in words, without tree structure) between an event trigger and an entity mention. Off-path information are in \textit{italics}. In these examples, the off-path words are prepositional words that help with the correct classification of the entity mention's argument role even though all the entities are of the same type, namely ``quantity'': 
\begin{enumerate}[topsep=3pt]
    \item \textbf{soared} \textit{by} \underline{1.350 million barrels} \\
   argument role - difference
   \item \textbf{soared} \textit{from} \underline{a mere 200 million barrels}: \\
   argument role - initial value
   \item  \textbf{soared} \textit{to} \underline{438.9 million barrels} \\
   argument role - final value
\end{enumerate}

\begin{figure*}[ht!]
\includegraphics[width=\textwidth]{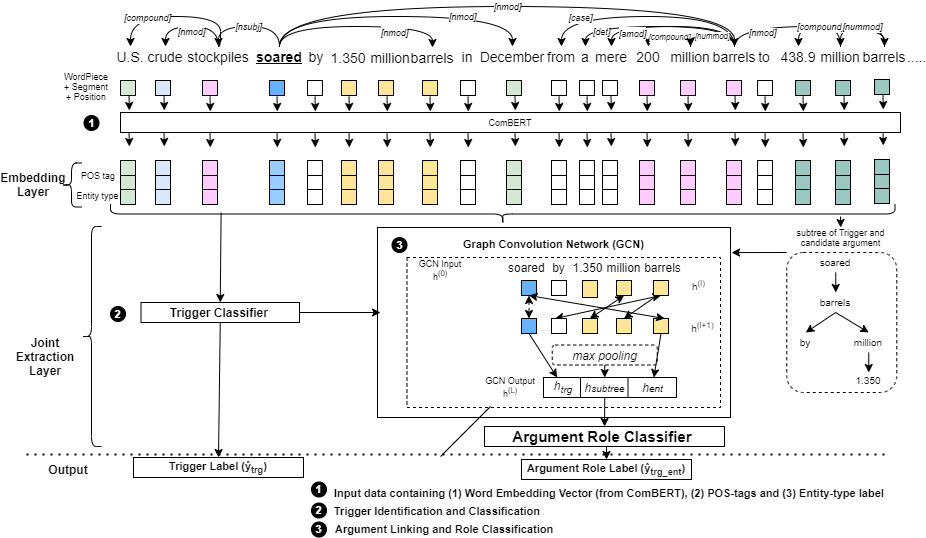}
\caption{Architecture of the framework}
\label{fig:architure}
\end{figure*}
\vspace{-0.7em}

\subsection{Domain-Adaptive Pre-training: ComBERT} \label{subsection:ComBERT}
Pre-trained language models such as BERT have been shown to produce SOTA results in many NLP tasks, including event extraction in the ACE2005 corpus \cite{yang2019exploring}. Instead of using BERT, we have decided to further pre-train BERT on a commodity news corpus, adapting the model to the finance and economics domains. The resulting model is referred to here as ComBERT. In the commodity news corpus, there are some commodity-specific polysemous words that can be better represented with further pre-training with in-domain data. Apart from the famous 'bank' example where the word could mean (1) financial institution or (2) terrain that is part of the river, there are some commodity-specific polysemous words in the commodity news corpus that can be better represented with further pre-training with in-domain data, for example: 
\begin{itemize}[topsep=3pt]
    \item \textbf{\texttt{stocks}}: (1) inventory and (2) shares
    \item \textbf{\texttt{tank}}: (1) storage vessel (noun), (2) market / price drop (verb)
\end{itemize}

The commodity news corpus is made up of about 20k news articles extracted from \href{https://www.investing.com/commodities/crude-oil-news}{https://www.investing.com/commodities/crude-oil-news}\footnote{the same source as the annotated dataset used here in event extraction.}, with publishing dates ranging from 2013 to 2019. We initialized ComBERT with \texttt{bert-base-cased}, one of the pre-trained BERT models provided by \cite{devlin-etal-2019-bert} and have the same model settings of transformer and pre-training hyperparameters as BERT. We have decided to use cased vocabulary \texttt{bert-base-cased} instead of uncased because the event extraction task involves extracting event arguments that are made up of named entities and nominal entities. Examples of named entities are countries, organizations, and specific commodities-related terms such as WTI, ICE, NYMEX, Brent and etc. Having a case-sensitive model yielded slightly better performance for the downstream event extraction task. 



The proposed solution is built on leveraging the power of a domain-adaptative pre-training model, ComBERT, to further fine-tune for event extraction with a smaller dataset and with fewer training steps. 

\subsection{Event Extraction}
Both of the sub-tasks within event extraction: (1) event trigger extraction and (2) event argument extraction are conceived as sentence-level multi-label classification tasks. The overall architecture is shown in Figure \ref{fig:architure}.
\subsection{Data Proprocessing}
\paragraph{Input.} The annotation files made public in \cite{lee2021annotated} were first converted from Brat Annotation standoff format (\textit{.ann} files) along with their corresponding news articles (\textit{.txt} files)  to \textit{json} format. Each sentence in the dataset was parsed using Stanford CoreNLP toolkit, including sentence splitting, tokenization, POS-tagging, NER-tagging, and dependency parsing to generate dependency parse trees. For input to the model, we adopt the "multi-channel" strategy (shown in ``1'' in Figure \ref{fig:architure}) by concatenating three components listed below.  Let \textit{W} = $ w_1, w_2,.....w_n $ be a sentence of length n where $ w_i $ is the \textit{i}-th token:
\begin{enumerate}[topsep=3pt]
    \item The word embedding vector of $w_i$: this is the feature representation from a word embedding of ComBERT. It is made up of WordPiece tokenization \cite{DBLP:journals/corr/WuSCLNMKCGMKSJL16} with [CLS] and [SEP]\footnote{[CLS], [SEP], [MASK] are special tokens of BERT. For experiments involving RoBERTa, Byte-Pair Encoding (BPE) tokenization and its special tokens are used.} are placed at the start and end of the sentence.
    \item The Part-of-Speech-tagging (POS-tagging) label embedding vector of $w_i$: This is generated by looking up the POS-tagging label embedding.
    \item The entity type label embedding vector of $w_i$: Similar to the POS-tagging label embedding vector of $w_i$, entity mentions in a sentence were annotated using BIO annotation schema, and the entity type labels were transformed to real-valued vectors by looking up the entity label embedding.
\end{enumerate}

\subsubsection{Trigger Candidate and Argument Joint-Extraction}
The experiments in this paper were conducted based on the joint-extraction approach where event trigger and arguments extraction are trained together. 
\paragraph{Event Trigger Extraction.} The event trigger extraction is setup as a token classification task, similar to that of Named Entity Recognition. The trigger classifier (shown as ``2'' in Figure \ref{fig:architure}) is a simple multi-layer perceptron (MLP) with a single hidden layer where each token within the input sentence is classified into one of 19 classes (18 event types and 'NONE' for non-event). All predicted event triggers are added to a list, $\widehat{y}_{trg}$ where $\widehat{y}_{trg}$ = $ t_1, t_2,.....t_j $. 
\paragraph{Event argument extraction.} For the second sub-task, the event argument extraction task is setup as a sequence classification task. Candidate arguments are selected from the pool of entity mentions within the sentence. Each candidate argument will be paired with a candidate trigger for argument role classification. The classifier will classify each trigger-entity pair into one of 20 classes (19 argument roles and 'NONE' for entities with no links to the candidate trigger). 

In the experiments, gold-standard entity mention \textit{E} were used, \textit{E} = $ e_1, e_2,...e_k $ where $ k $ is the number of the entity mentions in a sentence. With the list of predicted candidate triggers $\widehat{y}_{trg}$ with $j$ number of triggers, and the list of golden entity mentions $E$ with $k$ number of entity mentions, we pair each candidate trigger with an entity, resulting in $j$ $\times$ $k$ number of pairs. For the pair $t_x e_y$, the task is to classify the argument roles entity $e_y$ plays in event $t_x$.

\subsubsection{Graph Convolutional Networks over Dependency Tree}
A sentence's syntactic parse tree can be seen as a directed graph. Let $\CMcal{G} = \{\CMcal{V}, \CMcal{E}\}$ be the dependency parse tree for the sentence $w$ with 
$\CMcal{V}$ and $\CMcal{E}$ as the sets of nodes and edges of $\CMcal{G}$ respectively. $\CMcal{V}$ contains $n$ nodes corresponding to the $n$ tokens $w_1,w_2,....,w_n$ in $w$. Each edge ($v_i, v_j$) $\in$ $\CMcal{E}$ is directed from the head word $w_i$ to the dependent word $w_j$) with the Universal Dependency (UD) relation tags. 
Given a sentence's dependency parse tree with with $n$ nodes, we convert each tree into its corresponding $n \times n$ adjacency matrix \textbf{$A$} with the following modifications:
\begin{enumerate}[topsep=3pt]
    \item Treating the dependency graph as undirected, i.e $\forall i,j, A_{i,j} = A_{j,i}$, where $A_{i,j}$ = $A_{j, i}$ = 1 if there is a dependency edge between tokens $i$ and $j$;
    \item Adding self-loops to the each node in the graph, following \cite{DBLP:conf/iclr/KipfW17}: 
    $\widetilde{A} = A + I$ with $I$ being the $n \times n$ identity matrix
\end{enumerate}

Stacking a GCN layer $L$ times gives us a $L$-layer GCN where $L$ is a hyperparameter of the model. During graph convolution at each layer $l$, each node gathers and summarizes information from its connected nodes ($\widetilde{A}_{i,j}$ = 1) in the graph. We set $h^{(0)}$ as the input word vectors for an L-level GCN network and $h^{(L)}$ as the output word representations. The graph convolution operation of a single node, node $i$ at level $l$ of the GCN is as follows:
\begin{equation}
 h_i^{(l)} = \sigma (\sum_{j=1}^n \widetilde{A}_{ij}W^{(l)} h_j^{l-1} / d_i + b^{(l)})
    \label{eq:adfasdf}
\end{equation}
\noindent where  $h_i^{(l-1)}$ is the input vector, $h_i^{(l)}$ denotes the collective hidden representations, $W^{(l)}$ is the weight matrix, $b^{(l)}$ is a bias term, $\sigma$ is the sigmoid activation function and $d_i = \sum_{j=1}^n \widetilde{A}_{ij}$, is the number of arches in the resulting graph.

\subsubsection{Argument Role Classification with GCN}
This section describes the operations shown as ``3'' in Figure \ref{fig:architure}.
\paragraph{Encoding Trigger-Entity Pair.} Given a trigger-entity pair $t_xe_y$, we prune the dependency tree to obtain the contextual sub-tree between trigger $t_x$ and entity $e_y$ based on Algorithm \ref{algo}. The nodes of this sub-tree form the input word vectors $h^{(0)}$ to the L-layer GCN network.  

The subtree representation after $L$ times of graph convolution is obtained as follows:  
\begin{equation}
 h_{subtree} = f(h^{(L)}) = f(GCN(h^{(0)}))
    \label{eq:subtree}
\end{equation}

\noindent where $h^{(L)}$ is the output word representations produced by the L-layer GCN network and $f$ is a max-pooling function that maps the input to the subtree vector, $h_{subtree}$. Besides the subtree representation, we also obtained a representation $h_{trg}$ for trigger and $h_{ent}$ for entity:  
\begin{equation}
 h_{trg} = f(h_t^{(L)})    
, \\
h_{ent} = f(h_e^{(L)})  
\end{equation}
 
\noindent Besides max-pooling, we have also experimented with average-pooling and sum-pooling to obtain the final vector for all three vectors (substree, trigger and entity). All three vectors are then concatenated into a vector which is then propagated through a fully-connected layer to classify the argument role:
\begin{equation}
 \widehat{y}_{trg\_ent} = g(W_a[h_{subtree}; h_{trg};h_{ent}] + b_a)
    \label{eq:classrep}
\end{equation}
where $g$ is the \textit{softmax} operation to obtain a probability distribution over argument roles. $\widehat{y}_{trg\_ent}$ is the final output of the role the entity \textit{ent} plays in the event triggered by the trigger candidate \textit{trg}. 

\subsubsection{Loss Function for Joint-extraction}
Event trigger and arguments extraction are jointly trained by minimizing the cross-entropy loss function. 
\begin{equation}
 \mathcal{L}_{joint} = \mathcal{L}_{trg} + \beta (\mathcal{L}_{arg}) 
\end{equation}
where $\beta$ is the weightage placed on the loss of argument extraction task. In the experiment we use the value of 2, training the model with double the weightage on argument extraction.

\section{Experiment Setup} \label{sec:experiment}
\paragraph{Parameter settings.}
The data is split into 70\% for training and 30\% for testing. For all the experiments, the word embedding is of size 768 dimensions (same as \texttt{bert-base-cased}) while 50 dimensions for the other two embeddings - POS-tag embedding and entity-type embedding. For the GCN module, we use a two-layer GCN ($L$ = 2) with a batch size of 4. The model is trained using the Cross-entropy loss function and Adam optimizer.

 \begin{table*}[t]   
    \centering
    \begin{tabular}{ |p{2.5cm} |p{0.6cm} | p{0.6cm} | p{0.6cm} | p{0.6cm} | p{0.6cm}| p{0.6cm} |p{0.6cm} | p{0.6cm} | p{0.6cm} | p{0.6cm} | p{0.6cm} | p{0.5cm} |}  \hline
     & \multicolumn{3}{c|}{\textbf{Trigger}} & \multicolumn{3}{c|}{\textbf{Trigger}} & \multicolumn{3}{c|}{\textbf{Argument}} &  \multicolumn{3}{c|}{\textbf{Argument}} \\ 
    \textbf{Method} & \multicolumn{3}{c|}{\textbf{Identification (\%)}} & \multicolumn{3}{c|}{\textbf{Classification (\%)}} & \multicolumn{3}{c|}{\textbf{Identification (\%)}} &  \multicolumn{3}{c|}{\textbf{Role (\%)}} \\
    \cline{2-13}
      & P & R & F1 & P & R & F1 & P & R & F1 & P & R & F1 \\ \hline
    \small{Sequence representation \textbf{(A)}} & 
        \small{0.81} & \small{0.82} & \small{0.72} & 
        \small{0.75} & \small{0.71} & \small{0.70} & 
        \small{0.71} & \small{0.70} & \small{0.7} & 
        \small{0.60} & \small{0.62} & \small{0.62} \\ \hline
    \small{JMEE \textbf{(B)}} & \small{0.91} & \small{0.92} & \small{0.92} & \small{0.89} & \small{0.91} & \small{0.90} & \small{0.79} & \small{0.79} & \small{0.78} & \small{0.72} & \small{0.76} & \small{0.75} \\ \hline
    \small{GCN with full tree \textbf{(C)}} & 
        \small{0.92} & \small{0.97} & \small{0.95} & 
        \small{0.92} & \small{0.94} & \small{0.92} & 
        \small{0.80} & \small{0.81} & \small{0.81} & 
        \small{0.74} & \small{0.75} & \small{0.73} \\ \hline
    \small{GCN with LCA sub-tree \textbf{(D)}} & 
        \small{0.92} & \small{0.96} & \small{0.94} & 
        \small{0.93} & \small{0.95} & \small{0.94} & 
        \small{0.89} & \small{0.90} & \small{0.89} & 
        \small{0.87} & \small{0.86} & \small{0.85} \\ \hline 

    \small{GCN with contextual sub-tree \textbf{(E)}} & 
        \small{0.93} & \small{0.98} & \small{0.95} & 
        \small{0.91} & \small{0.97} & \small{0.93} & 
        \textbf{\small{0.92}} & \textbf{\small{0.91}} & \textbf{\small{0.92}} & 
        \textbf{\small{0.90}} & \textbf{\small{0.89}} & \textbf{\small{0.90}} \\ \hline
      \end{tabular}
    \caption{Comparing results (Precision, Recall, and F1 scores) across various methods with gold-standard entity mentions.}
    \vspace{-0.7em}
    \label{table:result1}
\end{table*}

\paragraph{Models Settings.}
The architecture and setup for models listed in Table~\ref{table:result1} are as follows: \\
(1) \textbf{Model A} - The embedding of trigger and candidate argument (from ComBERT) are concatenated and fed into a Bi-LSTM, which is then fed into a classifier with one fully connected (FC) layer. \\
(2) \textbf{Model B} - Jointly Multiple Events Extraction via Attention-based Graph Information Aggregation (JMEE) as presented in~\cite{liu-etal-2018-jointly}\footnote{This was developed for the ACE2005 dataset.}. \\
(3) \textbf{Model C} - GCN with Full Tree uses the full dependency tree, $h_{fulltree}$. The same convolution operations are done on $h_{fulltree}$ in the place of $h_{subtree}$. \\
(4) \textbf{Model D} - GCN with LCA sub-tree with shortest dependency path between trigger candidate and entity candidate. \\
(5) \textbf{Model E} -  GCN with contextual sub-tree, this setup is described in the Proposed Solution section. 

\section{Results and Analysis} \label{sec:results}
\subsection{Trigger Classification}
As shown in Table \ref{table:result1}, Trigger Identification and Trigger Classification achieve rather high F1 scores in all experiments regardless of whether full dependency tree or sub-tree, sequential or syntactic representation approach.

\begin{table*}[!ht]  
    \centering
    \begin{tabular}{ | p{3.5cm} | p{4.5cm} | c c c c c |}
    \hline
    & & \multicolumn{5}{c|}{\textbf{\small{Argument Role Classification F1 Score}}} \\
     \small{\textbf{Argument Roles}} & \small{\textbf{Entity Type}} & \small{\textbf{Model A}} & \small{\textbf{Model B}} & \small{\textbf{Model C}} & \small{\textbf{Model D}} & \small{\textbf{Model E}}\\ \hline
     
     \small{NONE} & - & \small{0.84} & \small{0.84} & \small{0.90} & \small{0.91} & \textbf{\small{0.94}}   \\ \hline
     
     \small{Attribute} & \small{Financial Attribute} & \small{0.40} & \small{0.65} & \small{0.79} & \small{0.75} & \textbf{\small{0.83}} \\ \hline
     
     \small{Item} & \small{Economic Item} & \small{0.64} & \small{0.85} & \textbf{\small{0.88}} & \small{0.85} & \textbf{\small{0.88}} \\ \hline
     
     \small{Final\_value} $\clubsuit$ & \small{Money / Production unit / Price unit / Percentage / Quantity} & \small{0.43} & \small{0.39} & \small{0.71} & \small{0.75} & \textbf{\small{0.79}} \\  \hline
     
     \small{Initial\_value} $\clubsuit$ & \small{Money / Production unit / Price unit / Percentage / Quantity} & \small{0.56} & \small{0.56} & \small{0.73} & \small{0.69} & \textbf{\small{0.77}} \\ \hline
     
     \small{Difference} $\clubsuit$ & \small{Money / Production unit / Price unit / Percentage / Quantity} & \small{0.58} & \small{0.69} & \small{0.84} & \textbf{\small{0.89}} & \textbf{\small{0.89}} \\ \hline
     
     \small{Reference\_point} $\diamondsuit$ & \small{Date} & \small{0.54} & \small{0.69} & \textbf{\small{0.80}} & \small{0.71} & \textbf{\small{0.80}} \\ \hline
     
     \small{Initial\_reference\_point} $\diamondsuit$ & \small{Date} & \small{0.40} & \small{0.63} & \small{0.63} & \small{0.60} & \textbf{\small{0.66}}  \\ \hline
     
     \small{Contract\_date} $\diamondsuit$ & \small{Date} & \small{0.52} & \small{0.54} & \small{0.70} & \small{0.66} & \textbf{\small{0.80}}  \\ \hline
     
     \small{Duration} & \small{Duration} & \small{0.55} & \small{0.55} & \small{0.75} & \small{0.82} & \textbf{\small{0.84}}  \\ \hline
     
     \small{Type} & \small{Location} & \small{0.52} & \small{0.59} & \small{0.70} & \small{0.68} & \textbf{\small{0.76}} \\ \hline
     
     \small{Imposer} $\spadesuit$ & \small{Country / State or province} & \small{0.71} & \small{0.69} & \textbf{\small{0.81}} & \small{0.79} & \textbf{\small{0.81}} \\ \hline
     
     \small{Imposee} $\spadesuit$ & \small{Country / State or province} & \small{0.50} & \small{0.49} & \small{0.60} & \textbf{\small{0.68}} & \textbf{\small{0.68}}  \\ \hline
 
      \small{Place} $\spadesuit$ & \small{Country / State or province} & \small{0.58} & \small{0.69} & \textbf{\small{0.74}} & \small{0.60} & \textbf{\small{0.74}} \\ \hline
      
     \small{Supplier\_consumer} $\spadesuit$ & \small{Country / State or provience / Nationality / Group} & \small{0.49} & \small{0.71} & \small{0.73} & \small{0.73} & \textbf{\small{0.79}} \\ \hline
     
     \small{Impacted\_countries} $\spadesuit$ & \small{Country} & \small{0.42} & \small{0.69} & \small{0.72} & \small{0.70} & \textbf{\small{0.76}}  \\ \hline

     \small{Participating\_countries} $\spadesuit$ & \small{Country} & \small{0.65} & \small{0.75} & \small{0.78} & \small{0.83} & \textbf{\small{0.89}} \\ \hline
     
     \small{Forecaster} & \small{Organization / Group} & \small{0.62} & \small{0.75} & \small{0.78} & \small{0.80} & \textbf{\small{0.82}} \\ \hline
     
     \small{Forecast} & \small{Forecast\_Target} & \small{0.61} & \small{0.61} & \small{0.83} & \small{0.67} & \textbf{\small{0.91}} \\ \hline
     
     \small{Situation} & \small{Phenomenon / Other acitivites} & \small{0.57} & \small{0.69} & \textbf{\small{0.73}} & \small{0.67} & \small{0.66} \\
     \hline 
      \end{tabular}
      \caption{F1-scores for each argument type.}
     \label{table:analysis_by_argument}
\end{table*}

\subsection{Argument Role Classification}
From the results shown in Table~\ref{table:result1}, it can be concluded that syntactic representation \textbf{(Model C, D, E)} of a sentence yields better event extraction results. The results of Model B and C are not as good as using sub-tree because the full dependency tree contains unnecessary and noisy information that is not helpful in argument role classification. As for \cite{liu-etal-2018-jointly} (Model B), it did not produce the best results because it was designed for capturing the association between multiple events within a sentence via the attention mechanism. The events in the commodity news dataset do not exhibit the same strong association as the events in ACE2005 dataset. Model D uses the LCA sub-tree that has only the ``bare minimum'' information, while Model E contains additional crucial context information that has proved to be useful in argument role classification. 

Table~\ref{table:analysis_by_argument} presents the breakdown of Argument Classification by Argument Types to fully provide evidence to the effectiveness of the proposed solution on the corpus, which exhibits the characteristic of arguments homogeneity. It is shown clearly that arguments of the same entity type, for example, \textit{Final\_value}, \textit{Initial\_value} and \textit{Difference} can be better differentiated and classified using a contextual sub-tree that contains the shortest path between an event trigger and its event argument as well as crucial off-path information.
Symbols ($\clubsuit$, $\diamondsuit$, $\spadesuit$) in Table~\ref{table:analysis_by_argument} indicate the grouping of arguments by entity type.

As for the characteristic of having multiple events in a sentence, the proposed solution is able to detect and classify the events as well as link arguments to their rightful event, as shown in both Table~\ref{table:result1} and Table~\ref{table:analysis_by_argument}. 

\subsection{Comparing Word Embedding and Pre-trained Language Models}
\textbf{Model E} in Table \ref{table:various_word_embedding} was further experimented using GloVe~\cite{pennington2014glove} and other pre-trained language models namely BERT~\cite{DBLP:conf/naacl/DevlinCLT19} and RoBERTa~\cite{liu2019roberta}. These were compared against ComBERT. 
\begin{table}[t!]   
    \centering
    \begin{tabular}{ |c | c | c| c |c | c | c| } \hline
    & \multicolumn{3}{c|}{\textbf{\small{Trigger}}} &   \multicolumn{3}{c|}{\small{\textbf{Argument}}} \\ 
    \textbf{Method}  & \multicolumn{3}{c|}{\small{\textbf{Classification (\%)}}} &   \multicolumn{3}{c|}{\small{\textbf{Role Class. (\%)}}} \\ \hline
    & \small{P} & \small{R} & \small{F1} & \small{P} & \small{R} & \small{F1} \\ \hline
    \small{GloVe} & 
        \small{0.67} & \small{0.70} & \small{0.68} & 
        \small{0.65} & \small{0.66} & \small{0.67} \\ 
        \hline
    \small{BERT} & 
        \textbf{\small{0.93}} & \small{0.93} & \small{0.93} & 
        \small{0.85} & \small{0.83} & \small{0.83} \\ 
        \hline
    \small{RoBERTa} & 
        \small{0.89} & \small{0.96} & \small{0.93} & 
        \small{0.86} & \small{0.85} & \small{0.83} \\ 
        \hline
    \small{\textbf{ComBERT}} & 
        \small{0.91} & \textbf{\small{0.97}} & \textbf{\small{0.93}} & 
        \textbf{\small{0.90}} & \textbf{\small{0.89}} & \textbf{\small{0.90}} \\ \hline
      \end{tabular}
      \caption{Comparing Word Embedding and Pre-trained Language Models for Model E}
      \vspace{-1em}
    \label{table:various_word_embedding}
\end{table}

From the results in Table \ref{table:various_word_embedding}, it is shown that ComBERT produced the best result, further proving that a contextualized token representation helps boost the performance of event extraction. ComBERT is used in all models listed in Table~\ref{table:analysis_by_argument}.

\section{Conclusion} \label{sec:conclusion}
\vspace{-0.2em}
This paper presents a graph-based deep learning framework to extract commodity-related events. This framework is tailored for the purpose of commodity news event extraction based on the new commodity news dataset introduced by \cite{lee2021annotated}. Our method addresses specific challenges exhibited by the characteristics of this dataset, in particular: (1) sentences containing lots of numerical information such as price, percentage of change, and dates, (2) entities of similar type playing distinctly different argument roles, and (3) the need for arguments extraction to disambiguate the identified events. The proposed solution uses a Graph Convolutional Network with a contextual sub-tree to extract events effectively. Experimental results demonstrate that the proposed solution outperforms existing solutions with higher F1 scores, particularly in argument role classification. With accurate event extraction from commodity news, the extracted information can be used for other downstream tasks such as learning event chains and event-event relations that can be further exploited for commodity price prediction. 

\clearpage
\bibliography{anthology,custom}
\bibliographystyle{acl_natbib}

\appendix
\onecolumn
\section{Commodity News Dataset} \label{app:Dataset}
\subsection{Event Types} 
There are 18 event types that can be grouped into four main categories: (1) macro-economic, (2) geo-political, (3) supply-demand-related, and (4) commodity price movement events. In-depth details about the event types such as the categories they belong to, description, and as well as more example sentences are found in \cite{lee2021annotated}.
\begin{table}[h!]   
    \centering \small
    \begin{tabular}{| p{0.23\linewidth} |p{0.70\linewidth} |}  \hline
    \textbf{Event type} & \textbf{Examples } \\ \hline
    1. Cause-movement-down-loss & 
    The pandemic has \textbf{zapped} demand to a level never seen before...\\ \hline
    2. Cause-movement-up-gain & 
    IEA tried to \textbf{boost} global oil demand by introducing.....\\ \hline
    3. Civil-unrest & 
    .....a fragile recovery in Libyan supply outweighed \textbf{fighting} in Iraq ......\\ \hline
    4. Crisis & 
     .....Ukraine declared an end to an oil \textbf{crisis} that has .....\\ \hline
    5. Embargo & 
    .....prepared to impose `` strong and swift '' economic \textbf{sanctions} on Venezuela...\\ \hline
    6. Geopolitical-tension & 
    ....despite geopolitical \textbf{war} in Iraq , Libya and Ukraine.\\ \hline
    7. Grow-strong & 
    .....as \textbf{strong} U.S. employment data.....\\ \hline
    8. Movement-down-loss & 
    ....further \textbf{decreases} in U.S. crude production.....\\ \hline
    9. Movement-flat & 
    U.S. crude is expected to \textbf{hold} around \$105 per barrel. \\ \hline
    10. Movement-up-gain &
    It expects consumption to \textbf{trend upward} by 1.05 million bpd. \\ \hline
    11. Negative-sentiment & 
    ....due to \textbf{concern} about softening demand growth and awash in crude.\\ \hline
    12. Oversupply & 
    ....the market is still working off the \textbf{gluts} built up.....\\ \hline
    13. Position-high & 
    Oil price remained close to four-year \textbf{highs}....\\ \hline
    14. Position-low & 
    Oil slipped more than 20\% to its \textbf{lowest level} in two years on 1980s...\\ \hline
    15. Prohibiting & 
    U.S. has \textbf{prohibit} the sale of oil from Venezuela.\\ \hline
    16. Shortage & 
    ......and there is no \textbf{shortfall} in supply , the minister added.\\ \hline
    17. Slow-weak & 
    U.S. employment data \textbf{contracts} with the euro zone....\\ \hline
    18. Trade-tensions & 
    ... escalating global \textbf{trade wars}, especially between the US and China. \\ \hline 
    \end{tabular}
    \caption{Event types with examples.}
    \label{table:EventDistribution}
    \vspace{-0.7em}
\end{table}

\subsection{Event Schema}
Each event has its own set of event arguments. Below is the complete list of arguments for the event \textbf{movement\_up\_gain}. The argument text of this table is populated using the example sentence presented in Figure \ref{fig:example1}. For the complete list of event schemas refer to \cite{lee2021annotated}.
 \begin{table}[h!]   
    \centering \small
    \begin{tabular}{ | p{0.20\linewidth} | p{0.52\linewidth} | p{0.20\linewidth} |}  \hline
    \textbf{Argument Role} & \textbf{Entity Type} & \textbf{Argument Text}\\ \hline
     Type & Nationality, Location &  \\ \hline
     Supplier\_consumer & Organization, Country, State\_or\_province, Group, Location & U.S. \\ \hline
     Reference\_point\_time & Date & Tuesday \\ \hline
     Initial\_reference\_point & Date & \\ \hline
     Final\_value & Percentage, Number, Money, Price\_unit, Production\_unit, Quantity & 438.9 million barrels \\ \hline
     Initial\_value & Percentage, Number, Money, Price\_unit, Production\_unit, Quantity &  200 million barrels \\ \hline
     Item & Commodity, Economic\_item & crude \\ \hline
     Attribute & Financial\_attribute & stockpiles\\ \hline
     Difference & Percentage, Number, Money, Production\_unit, Quantity & 1.350 million barrels\\ \hline
     Forecast & Forecast\_target &  \\ \hline
     Duration & Duration & \\ \hline
     Forecaster & Organization & \\ \hline 
    \end{tabular}
    \caption{Event Schema for the event \textbf{movement\_up\_gain}.}
\end{table}

\clearpage
\subsection{Argument Roles and Entity Types}
Table \ref{table:arg_ent_appendix} is a repeat of Table \ref{table:analysis_by_argument} showing the association of various entity types to each argument roles. The table below provides examples of entity mention for each argument roles.
\begin{table}[!ht]  
    \centering \small
    \begin{tabular}{ | p{0.18\linewidth} | p{0.25\linewidth} | p{0.50\linewidth} |} \hline
     \textbf{Argument Roles} & \textbf{Entity Type} & \textbf{Examples}\\ \hline
     1. NONE & - & -   \\ \hline
     2. Attribute & Financial Attribute & \textit{supply, demand, output, production, price, import, export} \\ \hline
     3. Item & Economic Item & \textit{economy, economic growth, market, economic outlook, employment data, currency, commodity-oil} \\ \hline
     4. Final\_value & Money / Production unit / Price unit / Percentage / Quantity & \textit{\$60, USD 50, 170,000 bpd, 400,000 barrels per day, 1 million barrels, \$40 per barrel, USD58 per barrel, 0.5\%}\\  \hline
     5. Initial\_value & Money / Production unit / Price unit / Percentage / Quantity & \texttt{--same as above--}\\ \hline
     6. Difference & Money / Production unit / Price unit / Percentage / Quantity &  \texttt{--same as above--}\\ \hline
     7. Reference\_point & Date & \textit{1998, Wednesday, Jan. 30, the final quarter of 1991, the end of this year} \\ \hline
     8. Initial\_reference\_point & Date &  \texttt{--same as above--}\\ \hline
     9. Contract\_date & Date & \texttt{--same as above--}\\ \hline
     10. Duration & Duration & \textit{two years, three-week, 5-1/2-year, multiyear, another six months} \\ \hline
     11. Type & Location &  \textit{global, world, domestic}\\ \hline
     12. Imposer & Country / State or province &  \textit{China, Iran, Iraq, Washington, Moscow, Cushing, North America, Europe}\\ \hline
     13. Imposee & Country / State or province & \texttt{--same as above--} \\ \hline
     14. Place & Country / State or province &  \texttt{--same as above--}\\ \hline
     15. Supplier\_consumer & Country / State or provience /  Nationality / Group & \texttt{--same as above--} + \textit{OPEC, non-OPEC countries, American, Russian} \\ \hline
     16. Impacted\_countries & Country &  \textit{China, U.S. Russia, Iran, Iraq} \\ \hline
     17. Participating\_countries & Country & \texttt{--same as above--} \\ \hline
     18. Forecaster & Organization / Group & \textit{OPEC, European Union, U.S. Energy Information Administration} \\ \hline
     19. Forecast & Forecast\_Target &  \textit{forecast, target, estimate, projection, bets} \\ \hline
     20. Situation & Phenomenon  / Other activities&  \texttt{free text}\\
     \hline 
      \end{tabular}
      \caption{List of Event Argument Roles and their corresponding entity types.}
     \label{table:arg_ent_appendix}
\end{table}

\end{document}